\documentclass{article}
\usepackage{spconf,amsmath,graphicx}

\usepackage[utf8]{inputenc} 
\usepackage[T1]{fontenc}    
\usepackage{hyperref}       
\usepackage{url}            
\usepackage{booktabs}       
\usepackage{amsfonts}       
\usepackage{nicefrac}       
\usepackage{microtype}      
\usepackage{amsmath}
\usepackage{amsfonts}
\usepackage{amssymb}

\usepackage{adjustbox}  
\usepackage{color} 

\definecolor{cardinal}{rgb}{0.77, 0.12, 0.23}
\definecolor{officegreen}{rgb}{0.0, 0.5, 0.0}
	\definecolor{lightbrown}{rgb}{0.71, 0.4, 0.11}

\DeclareMathOperator*{\argmax}{arg\,max}

\title{Interpretable Concept-based Prototypical Networks for Few-Shot Learning}
\name{Mohammad Reza Zarei, Majid Komeili \thanks{Parts of this work was supported by NSERC and ComputeCanada.}}
\address{School of Computer Science, Carleton University, Canada\\
mohammadrezazarei@cmail.carleton.ca, majid.komeili@carleton.ca}
\begin{document}
%
\maketitle
\begin{abstract}
Few-shot learning aims at recognizing new instances from classes with limited samples. This challenging task is usually alleviated by performing meta-learning on similar tasks. However, the resulting models are black-boxes. There has been growing concerns about deploying black-box machine learning models and FSL is not an exception in this regard. In this paper, we propose a method for FSL based on a set of human-interpretable concepts. It constructs a set of metric spaces associated with the concepts and classifies samples of novel classes by aggregating concept-specific decisions. The proposed method does not require concept annotations for query samples. This interpretable method achieved results on a par with six previously state-of-the-art black-box FSL methods on the CUB fine-grained bird classification dataset.


\end{abstract}
\section{Introduction}
Category recognition is one of the fundamental tasks in computer vision, an area where neural networks have had a great success. However, they usually require a relatively large amount of labeled training data for each class. This may limit their application in scenarios where training data are scarce. To address this issue, few-shot learning (FSL) has been considered where the model has to generalize to novel classes with only a few instances. These classes are disjoint from the training classes where sufficient data is available for them during the training stage. 

Few-Shot Learning methods usually mimic the few-shot task by utilizing sampled mini-batches called episodes during the training stage. In each episode, a set of C\ classes are randomly selected from training classes. For each of these classes, K labeled instances are sampled to act as the support set, and a subset of the remainder serves as the query set \cite{Sung_2018_CVPR}. This setting is referred to as “C-way K-shot”. By using episodic learning, FSL attempts to improve the model’s generalization ability in tasks with few instances and transfer the learned knowledge of the model to few-shot learning problem for novel classes. This paradigm that is utilized in FSL models is referred to as meta-learning.

Recent meta-learning models can be roughly grouped into two categories. The first one, known as Optimization-based methods \cite{DBLP:journals/corr/abs-1807-05960,antoniou2018how} aims to fine-tune the learned model on the target task. The second category focuses on learning a metric space shared between source tasks \cite{Sung_2018_CVPR,Proto}. This space will be used to solve the target task by nearest neighbor search or learning a simple linear classifier on top of the model \cite{cao2021concept}. 

Although FSL models have been able to achieve remarkable performance in terms of accuracy in recent years, they are black-box models. There has been a growing concern about use of black-box models in real-world and FSL is not an exception. In general, Interpretability maybe be accomplished by applying post-hoc analysis methods on black-box models that are already trained, or alternatively we can create models that are interpretable by design. In the area of FSL, the main stream approaches have been posthoc \cite{Kang_2021_ICCV,8794911} and to the best of our knowledge, research on FSL methods that are inherently interpretable has rarely been conducted.

Inspired by \cite{Proto}, recently Cao and et. al proposed COMET \cite{cao2021concept}, a method for FSL along human-interpretable concept dimensions. When human tries to learn new bird species, they are already equipped with some structured, reusable concepts such as wing, beak, legs and feather that help efficiently adapt to the new task and also explain their decisions in terms of such concepts. 
COMET learns an embedding space for each concept by masking areas related to that concept. This method learns one metric space over each concept and the final decision is based on averaging decisions of different spaces.
\begin{figure*}[ht]
  \centering 
  \includegraphics[width=0.9\textwidth]{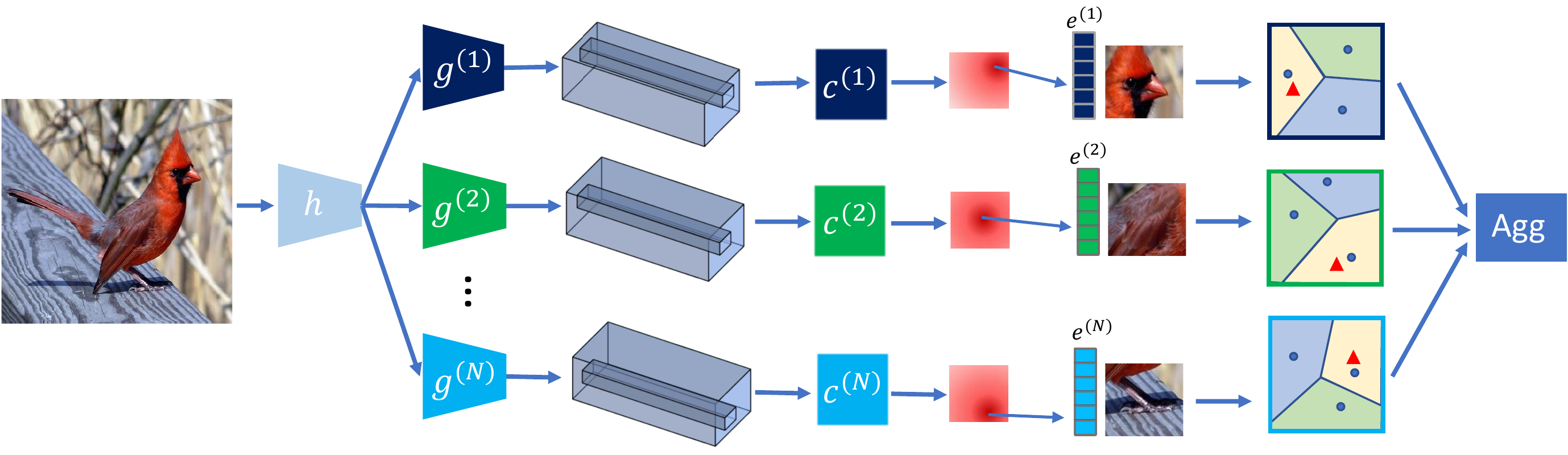}
  \caption{ Pipeline for performing FSL with InCoPoN.}
\end{figure*}

Although the aforementioned method is able to provide interpretable decisions, the concept annotations have to be given as prior knowledge not only for the instances of base tasks, but also for the instances of target tasks including the test samples. However, such concept annotations may not be readily available at test time.
While assuming the availability of annotations and labels for training samples, including the few training samples of novel classes is a typical assumption in supervised learning, we argue that extending such assumption, even partially, to test samples may limit the application of the resulting model in many real-world scenarios. 
Besides, COMET \cite{cao2021concept} does not properly handle commonalities and differences across concepts. They examined two extreme cases: training individual networks for concepts, and training a single network. However, the former ignore the commonalities among concepts and the latter ignores the differences between concepts by forcing the embeddings to be picked from the same feature map. 
Another drawback of COMET \cite{cao2021concept} is that at the aggregation step, all concepts have equal votes even those that are weakly present or are completely absent in the input. 

In this paper, we present Interpretable Concept-based Prototypical Networks (InCoPoN) to perform FSL based on a set of human-interpretable concepts. InCoPoN learns a set of concept-specific metric spaces, and extracts concept-specific embeddings for query samples and aggregates the resulting concept-specific decisions to make a final decision. 
The closest work to the proposed method is \cite{cao2021concept} where the three drawbacks described above are addressed.
The contributions of this paper are as follows: 
1) We propose an inherently interpretable method that unlike COMET \cite{cao2021concept}, does not need concept annotations for the test samples. The proposed method learns to infer them from the test samples. 
2) A multi-tasking approach that includes a shared back-bone network for capturing the commonalities among concepts followed by individual heads to capture the differences between concepts.
3) For aggregating concept ebmeddings, we propose an adaptive approach where for each sample, it emphasizes more on the concepts that are present in the sample. 
Through experiments, we show that the proposed interpretable method performs on a par with several previously state-of-the-art black-box FSL methods on the fine-grained bird classification task using CUB dataset \cite{WahCUB_200_2011} which is a widely used and yet challenging dataset due to the presence of highly similar classes. Moreover, through a detailed ablation study we demonstrate the effectiveness of the second and third contributions over some baseline methods.




\section{Proposed Method}
\label{proposed}
Given a labeled dataset $\mathcal{B}  = {\{(x_i,y_i)\}}_{i=1}^{N_b}$ for base classes $\mathcal{Y}_b$, and a labeled support set $\mathcal{S}={\{(x_i,y_i)\}}_{i=1}^{N_s}$ for novel classes ${\mathcal{Y}_n}$ where ${\mathcal{Y}_b} \cap {\mathcal{Y}_n} = {\emptyset}$, FSL aims to predict labels of a query set $\mathcal{Q} = {\{(x_i,y_i)\}}_{i=1}^{N_q}$ which also belongs to ${\mathcal{Y}_n}$. Samples in $\mathcal{B}$, $\mathcal{S}$ are annotated for $N$ common high-level concepts $\mathcal{C} = {\{c^{(k)}\}}_{k=1}^{N}$. Part-based annotations are associated with, for example, meaningful body parts of birds such as beak, belly and wings. For each concept, when is present in an image, only its location would  suffice. A bounding box or pixel-based segmentation is not required.


The proposed method consists of three main components: 1) Concept learners that provide concept-specific feature maps; 2) Concept detectors that predict location of concepts in the corresponding feature maps. For each concept, the outputs of this part is a probability score for presence of the concept along with the corresponding concept embedding vector; 3) Aggregation module that makes the final decision. These components will be described in the following.


\subsection{Concept learners with shared layers and concept-specific heads}
\label{concept_learners}
For each concept $c^{(j)}$, one embedding network $f_{\theta}^{(j)}: \mathcal{X}\rightarrow\mathcal{E}_{j}$, is learned where $\mathcal{X}$ is the image space and $\mathcal{E}_{j}$ is the embedding space for concept $j$. 
Intuitively, each embedding space $\mathcal{E}_{j}$ is learned to cluster samples around the prototype of their corresponding class only based on the concept $c^{(j)}$. This can be achieved by masking out non-concept regions of the input samples to ensure the concept learner sees only the concept-related parts of the input samples during training. Alternatively, the entire image can be used without masking to get an intermediate feature map $M^{h\times w \times c}$ and from there a feature vector $e_{x_i}^{(j)}$ corresponding to the center location of the concept in the input image can be picked. $h$, $w$ and $c$ denotes height, width and number of channels in the feature map. 
Note that this can be done because the locality is preserved when applying filters. 
Following \cite{cao2021concept}, we use the second approach. $e_{x_i}^{(j)}$ is a vector of the length $c$ and represents the concept embedding for sample $x_i$.


In \cite{cao2021concept}, two different designs were considered for concept learners. The first one learns one totally separate network $f^{(j)}$ for each concept that results in ignoring commonalities along different concepts. In the second design, a shared network $f$ is trained for all concepts, ignoring the differences between concepts. To consider both commonalities and differences, we design the concept learners to share weights in early layers and have their own concept-specific heads. Therefore, the concept learner $f^{(j)}$ is substituted with $g^{(j)}\circ h$ where $h$ is the network with shared parameters and $g^{(j)}$ is the network head for concept $j$.


The concept learners are trained on the images of base classes $\mathcal{Y}_b$ using episodic learning to mimic the few-shot classification setting. Using each concept learner $j$, one concept-specific prototype $P_y^{(j)}$ is calculated for class $y$ by averaging the concept embeddings of support set:
\begin{equation}
    P_y^{(j)} = \frac{1}{|S_y|} \sum \limits_{x_i \in {S_y}} e_{x_i}^{(j)}
\end{equation}
where $e_{x_i}^{(j)}$ is the concept embedding feature vector picked from $g^{(j)}\circ h(x_i)$ and $|S_y|$ is the number of images in the support set of class $y$.

For a query image $x_q$ in an arbitrary training episode, the concept embeddings are extracted using concept learners. Then by calculating an aggregated distance to concept-specific prototypes of different classes, the class of $x_q$ is determined. Specifically, to calculate the aggregated distance from the concept-specific prototypes of class $y$, the distance of each concept embedding $e_{x_q}^{(j)}$ from the concept prototype $P_y^{j}$ is calculated. Finally, the distances across all concepts are summed to calculate the probability of assigning $x_q$ to class $y$ as:
\begin{equation}
    p(y|x_q) = 
    \frac{exp(-\sum_{j \in \mathcal{C}} d(e_{x_q}^{(j)},P_y^{(j)}))}
    {\sum_{y'}exp(-\sum_{j \in \mathcal{C}} d(e_{x_q}^{(j)},P_{y'}^{(j)}))}
\end{equation}
$h$ and $g^{(j)}$ are trained using the negative log-likelihood $L = - \log p(y_{x_q}|x_q)$ of true class in an episodic training setting using the images and concept location of the base classes. 

\subsection{Predicting concept locations for query samples}
Similar to the simulated episodes during training concept learners, we can perform few-shot classification in target space. However, since the concept locations are not available for query images, in the following, we will present an approach to predict them.

To detect the location of concept feature vector in the last feature map of concept learner, one binary classifier is trained on top of each learned concept embedding network using the concepts of base classes. Specifically, on top of the embedding network of concept $j$, a binary classifier $c^{(j)}$ is trained using binary cross-entropy loss to detect $e_{x_i}^{(j)}$ from other feature vectors in the last feature map of the concept learner. To train the classifier, the feature vectors corresponding to the center of the concept in the input images are presented to the classifier as positive instances. Feature vectors in other spatial locations of the final feature map are provided to the model as negative instances. 

To detect the feature vector of concept $j$ for an arbitrary image $x_a$, the image is fed to the network $g^{(j)}\circ h$ and in the final feature map $M$, each feature vector along the channel dimension is provided to the binary classifier $c^{(j)}$ and the feature vector with the highest probability is selected as the class embedding $e_{x_a}^{(j)}$.

As the number of negative instances is considerably more than the positive ones, the classes are weighted in the cross-entropy loss to alleviate  the adverse effect of the imbalance data. 

\subsection{Aggregation module}
\label{sub_setting1}
To perform few-shot recognition in target space, concept-specific prototypes for each class are computed using the procedure described in Section \ref{concept_learners} and for each query image $x_q$, the concept embeddings $\{e_{x_q}^{(j)}\}_{j \in \mathcal{C}}$ are detected from the final feature maps of concept learners using the trained concept-specific classifiers described in the previous subsection.

Finally, class of the query image $x_q$ is determined using Eq. (\ref{eq_classification}) by measuring the accumulated distance of its concept embeddings to concept-specific prototypes of each class. This distance is indicated by ${D_y}({x_q})$ and formulated in Eq. (\ref{eq_dis}). 
\begin{equation}
\label{eq_classification}
    \hat{y} = \argmax_{y \in \mathcal{Y}_n} 
    \frac{exp(-{{D_y}({x_q})})}
    {\sum_{y'}exp(-{D_{y'}}({x_q}))}
\end{equation}

\begin{equation}
\label{eq_dis}
    {D_y}({x_q}) =  \frac{\sum_{j \in \mathcal{C}} {w_{x_q}^{(j)}}{ d(e_{x_q}^{(j)},P_y^{(j)})}}{\sum_{j \in \mathcal{C}} {w_{x_q}^{(j)}}}
\end{equation}
 where $w_{x_q}^{(j)}$ is the inverse of the probability score obtained from the binary classifier $c^{(j)}$ for the selected feature vector $e_{x_q}^{(j)}$. Therefore, concept embeddings with higher probabilities will have a higher impact on the final classification decision and likewise concepts that are not present in the query sample will have a lower impact.

\section{Experiments and Evaluations}
\label{eval}
\subsection{Dataset and experimental settings}
We evaluate InCoPoN on Caltech-UCSD Birds-200-2011 (CUB) \cite{WahCUB_200_2011} dataset. This is a fine-grained bird classification dataset consisting of 11,788 images from 200 different categories with a total number of 15 parts/concepts locations. 
We follow the protocol provided in \cite{chen2018a} for splitting the dataset. The models are evaluated on the widely used 5-way setting. Specifically, in each episode, 5 classes are sampled randomly where k samples are provided for each class as support set to form the k-shot classification task. The query set contains 16 samples from the classes of the support set. The best model is chosen based on the accuracy on the validation set. For testing, 600 episodes are sampled randomly from novel classes and the mean accuracy and standard deviation are reported for these 600 episodes.

The FSL widely used backbone network Conv-4 \cite{10.5555/3045118.3045167} with an input size of 84 $\times$ 84 is adopted for concept learners. The first three blocks of this network are shared among different concept learners and the last block is the head specific to each concept.  Moreover, each concept-specific binary classifier is a two-layer MLP with 64 neurons in the hidden layer. Finally,  Euclidean distance is employed to measure the distance between concept embeddings and prototypes. 
Similar to \cite{cao2021concept}, standard data augmentation including random crop, rotation, horizontal flipping and color jittering is performed. Finally, concept learners are trained using Adam optimizer with a learning rate of $10^{-3}$. 

\subsection{Performance comparison}
We compare InCoPoN with six previously state-of-the-art FSL methods as shown in Table \ref{tab_comparison}. It can be seen that for both 5-way 5-shot and 5-way 1-shot settings, InCoPoN achieved results on a par with the black-box FSL models and yet provides interpretability through learning to learn along human-friendly concepts.
%
%
On 5-way 5-shot setting, our method is able to achieve an average accuracy of 78.6\% which outperforms previously state-of-the-art methods ProtoNet \cite{Proto}, MAML \cite{pmlr-v70-finn17a}, MatchingNet \cite{NIPS2016_90e13578}, and is only slightly behind the MetaOptNet \cite{NIPS2016_90e13578} and Baseline++ \cite{chen2018a} (1\% and 1.6\%). On 5-way 1-shot setting, our method records 57.9\% in terms of average accuracy. The top performer is MetaOptNet with an average accuracy of 62.2\%. The reason for achieving somewhat less competitive results on 1-shot setting could be attributed to the fact that not all concepts are available for each image in CUB dataset, and it is more likely that a specific concept has no representation for a class in 1-shot setting since each class is represented with just one support image. In that case, global average pooling of the final feature map in the model is used as the features for that missing concept.
COMET \cite{cao2021concept} achieved 85.3\% and  67.9\% on 5-shot and 1-shot settings respectively but it should be noted that because it gains from the additional concept annotations for the test samples, a direct comparison may not be fair.

\begin{table}
\caption{Results of 5-way 5-shot and 5-way 1-shot on CUB dataset. Average accuracy and standard deviation over 600 randomly sampled episodes are reported.}
\label{tab_comparison}
\centering
\begin{tabular}{lll}
\hline
Method & 5-way 5-shot  & 5-way 1-shot  \\
\hline
Baseline++ \cite{chen2018a} & 80.2 $\pm$ 0.6 & 61.4 $\pm$ 1.0 \\
MatchingNet \cite{NIPS2016_90e13578} & 75.9 $\pm$ 0.6 & 61.0 $\pm$ 0.9 \\
MAML \cite{pmlr-v70-finn17a} &  74.4 $\pm$ 0.8 & 52.8 $\pm$ 1.0 \\
RelationNet \cite{Sung_2018_CVPR} & 78.6 $\pm$ 0.7 & 62.1 $\pm$ 1.0 \\
MetaOptNet \cite{Lee_2019_CVPR} & 79.6 $\pm$ 0.6 & 62.2 $\pm$ 1.0 \\
ProtoNet \cite{Proto} & 76.1 $\pm$ 0.7 & 57.1 $\pm$ 1.0 \\

\hline
InCoPoN & 78.6 $\pm$ 0.7  & 57.9 $\pm$ 0.9  \\
\hline

\end{tabular}
\end{table}

\subsection{Effect of using probability scores as weights}
In this section we compare the proposed method with a baseline model that considers equal weights for all concept embeddings. The results are shown in Table \ref{tab_effect}.
It can be seen that the proposed aggregation module improves the performance in both 5-way 5-shot and 5-way 1-shot settings.

\begin{table}
\caption{Comparison between the proposed method and the baseline model with equal weights for all concepts.}
\label{tab_effect}
\centering
\begin{adjustbox}{width={0.45\textwidth},totalheight={\textheight},keepaspectratio}
\begin{tabular}{lll}
\hline
Method & 5-way 5-shot  & 5-way 1-shot  \\

\hline
InCoPoN with equal weights & 77.2 $\pm$ 0.7  & 57.6 $\pm$ 0.9 \\
InCoPoN with probability scores as weights & 78.6 $\pm$ 0.7  & 57.9 $\pm$ 0.9  \\
\hline
\end{tabular}
\end{adjustbox}
\end{table}

\subsection{The impact of backbone network design}

To evaluate our new design for the backbone network, a comparison is performed between COMET with its original designs and our new design as shown in Table \ref{tab_design}
%
%
While COMET with original designs achieved the same performance of 85.3\% in terms of average accuracy in 5-way 5-shot setting, the new design improved the performance by approximately 2\%. This improvement is even higher in 5-way 1-shot setting achieving 3.97\% higher average accuracy.

\begin{table} [h]
\caption{Comparison between COMET with our design and the original designs of backbone network.}
\label{tab_design}
\centering
\begin{adjustbox}{width={0.45\textwidth},totalheight={\textheight},keepaspectratio}
\begin{tabular}{lll}
\hline
Method & 5-way 5-shot  & 5-way 1-shot  \\
\hline

COMET shared w & 85.3 $\pm$ 0.5 & 67.9 $\pm$ 0.9 \\
COMET  with distinct networks & 85.3 $\pm$ 0.5 & 67.9 $\pm$ 0.9 \\
\hline
COMET ours  & 87.25 $\pm$ 0.46 & 71.87 $\pm$ 0.92 \\
\hline

\end{tabular}
\end{adjustbox}
\end{table}

\section{Conclusion}
\label{conc}
In this paper, an interpretable few-shot learning model was proposed. Although it decomposes the decision space into multiple metric spaces associated with human-interpretable concepts, it does not require concept annotations for test samples. The process of learning multiple metric spaces is efficiently modeled as a multi-tasking problem. The results are aggregated by considering the degree that each concept is present in the input. Finally, The proposed interpretable method achieved competitive average accuracy in the range of six previously state-of-the-art black-box FSL methods.


\bibliographystyle{IEEEbib}
\bibliography{Main}

\end{document}